\documentclass[a4paper]{article}
\usepackage{INTERSPEECH2014,amssymb,amsmath,epsfig}
\usepackage{times}
\usepackage{graphicx}
\usepackage{algorithm}
\usepackage{algpseudocode}
\usepackage{mypackage}
\ninept
\def\reg{{\rm\ooalign{\hfil
     \raise.07ex\hbox{\scriptsize R}\hfil\crcr\mathhexbox20D}}}

\title{Manifold-Kernels Comparison in MKPLS for Visual Speech Recognition}

\makeatletter
\def\name#1{\gdef\@name{#1\\}}
\makeatother \name{{\em Amr Bakry, Ahmed Elgammal}}

\address{Rutgers Univesity \\
{\small \tt amrbakry@cs.rutgers.edu, elgammal@cs.rutgers.edu}}

\begin{document}
\maketitle
\begin{abstract}

Speech recognition is a challenging problem. Therefore, considering the visual information besides the acoustic stream, is essential for improving the recognition accuracy in real-life un-constraint situations. 
MKPLS is shown to be successful framework for solving visual speech recognition. However, MKPLS lacks in-deep analysis to leverage its power. %needs a more discussion to tackle 
This work is intended to study two core building blocks in the pipeline of MKPLS: manifold parameterization and manifold kernel. For manifold parameterization, we study the effect of changing the number of centers and the regularization factor in order to find the best parameters of this step. %We also study different alternative to reason about the kernel choice in this framework.
%One common approach is to model the visual recognition as nonlinear optimization problem. 
Quantifying the similarity between visual units is challenging step for computing the kernel. %for solving this problem.
Therefore, we explore three kernel categories: matrix-based, curve-based and subspace-based kernels. Intuition behind each kernel choice is provided and quantitative comparison among them is conducted. %Embedding the visual units on a manifold and using manifold kernels is one way to measure these distances. 
%This work is intended to evaluate the performance of several manifold kernels for solving the problem of visual speech recognition. 
In other words, this study is intended to reason about the kernel choice for VSR. We compare different kernels using MKPLS framework.%, since it gives convenient way to explore different kernels. 
We use two public datasets: OuluVs and AvLetters databases. 

\end{abstract}

%%%%%%%%% BODY TEXT
\section{Introduction}
Audio visual speech recognition (AVSR) has been investigated intensively in the last few decades \cite{Potamianos2004}. Specially after bimodal fusion of audio and visual stimuli in perceiving speech has been demonstrated by the \emph{McGurk} effect \cite{McGurk1976}. For example, when the spoken sound /ga/ is seen as /ba/, most people perceive the sound as /da/ \cite{McGurk1976}. More specifically, with the advances in computer vision, visual speech recognition (VSR), also called lipreading, have attracted research attention \cite{Shiell2008}. VSR systems gain importance with the need for controlling machines verbally in noisy acoustic environment. An example of that is the car environment, where the noise (\eg from motor and radio) makes it very difficult for audio speech recognition. %controlling the car acoustically.
Another potential example is to control robots in outer space where there is no media for audio transmission.

Mapping between phonemes and visemes \footnote{Viseme is the visual phoneme. It is defined as the smallest discriminative unit for visual speech} tends to be many-to-one,i.e, the same viseme can appear for many different phonemes. This shows that visual information solely might not be enough for achieving the speech recognition task. As a result, VSR is challenging problem, specially, when using information only from plan marker-less and real-life images.

On the other hand, speaker identification is tightly coupled problem with speech recognition \cite{Luettin1996, Sanderson2004, Shiell2008}. Speaker identification is defined as the ability to identify the speaker within a group of users from solely speech related features, like voice or mouth motion. %Speaker identification is related to several research fields such as automatic access control, biometrics, and personal privacy issues.

The appearance is not the ideal features to be used for solving the lipreading problem, while the dynamics in the utterance video attracts the researchers. Graphical models have been used extensively in VSR and AVSR. One technique that can be used to extract the dynamics in the video is 
%is the best  videothe mouth area is the best Several techniques have been adopted to extact the Two main approaches are commonly used to capture the in VSR literature: a 
Hidden Markov Model (HMM).
%based approach and classifier based approach. %extracting visual descriptor approach. 
%In the HMM approach, after choosing suitable descriptor for the visual unit (usually visemes) corresponding to every node, this descriptor employs as observations for the model. Then HMM model is trained using Baum-Welch algorithm for 
HMM encodes the stochastic temporal relationship between sequence of observations~\cite{Rabiner1989}.  
In \cite{Matthews2002-avletters}, HMM was used for encoding the visual dynamics of speech using Active Shape Model (ASM) and Active Appearance Model (AAM). A more general graphical models, Dynamic Bayesian Network(DBN) model, has been used in \cite{Saenko2005} with different visual articulation units called articulatory features.
%Consequently, the Viterbi algorithm \cite{Rabiner1989} is used for classification. The classifier based approach is based on extracting a single feature vector for the whole clip of uttered phrase (usually single word, or short sentence), and train a classifier (usually SVM) based on that \cite{Zhao2009,Fu2007}. 
%Encoding the dynamics of speech video as a descriptor has a long history within lipreading research. Graphical models have been used extensively in VSR and AVSR. 
Graph embedding has been used in \cite{Zhou2010} to model the temporal relationship between frames, then the graph is used later, in \cite{Zhou2011}, for estimating a curve that represent the dynamics in video. Graphical based methods try to capture the smooth temporal changes between the used visual units, but they may loose some visual information that may be crucial for discriminating small speech chunks like single letter utterance.

On the other hand, the work in \cite{Zhao2009} is based on extracting a single spatio-temporal feature vector for representing the visual and temporal information for the whole speech video. In \cite{Shaikh2010} optical flow was used for extracting the whole word features. These two approaches outperform in the case of small size videos but it might be sensitive to frame outliers.
%In our method, we care about smoothness, since we extract the geometric deformation of the lip-moving manifold and at the same time use all the appearance information for learning a parameterization for this manifold. We test our model on two databases, one contains small clip (AVLetters) and the other database contains slightly longer clips (OuluVs).
%As the best of our knowledge, we are the first to use homeomorphic manifold analysis and KPLS in the field of visual speech recognition.
Manifold-KPLS (MKPLS) framework is proposed in \cite{Bakry2013}. %  is adopted in this work for VSR. % belongs to the second category.
It finds a concise low-dimensional embedding for each visual unit. MKPLS has three phases: manifold parameterization, manifold latent space embedding and manifold classification. It uses Kernel-based supervised dimensionality reduction technique, namely, Kernel-PLS (KPLS) \cite{Rosipal2002:KPLS}. 

Choice of the used kernel in KPLS is very critical to achieve the best performance. This part has not been discussed in details in \cite{Bakry2013}. In this work, we explore and discuss several similarity measure for visual units. In MKPLS, visual units  are represented by their manifold parameterizations.
Even though kernel discussion is done from the point of view of MKPLS. Our findings can be generalized to be applied for other frameworks.
We apply MKPLS for AVLetters \cite{Matthews2002-avletters} and 
OuluVs \cite{Zhao2009}, and the experimental results contrast the performance variation between the kernels. 

After this introduction, the problem statement, mathematical modeling and the %Section~\ref{sec:problemDef}. The evaluation 
MKPLS framework are presented in Section~\ref{sec:framework}. Then, the used kernels are presented in-details in Section~\ref{sec:kernels}. Finally, the experimental results and the applied databases are listed in Section~\ref{sec:experments}.
%%%%%%%%%%%%%%%%%%%%%%%%%%%%%%%%%%%%%%%%
\section{Manifold-KPLS}
\subsection{Problem Definition}
\label{sec:problemDef}
Given set of visual units, we need to recognize new test unit, and infer the identity of the speaker. The visual unit could be viseme, word or a complete sentence. For each training visual unit is assigned to specific speech class and specific speaker. %  to group them semantically by human perceiption of them. 
Both the training and testing visual units are represented by sequence of images (frames) extracted from the speech video. Each frame exposes only the mouth area of the speaker. % The frames are restricted to mouth area for the biometric and privacy reasons.

\begin{figure*}[!tbh]
	%\centering{
	\includegraphics[width=1.0\textwidth]{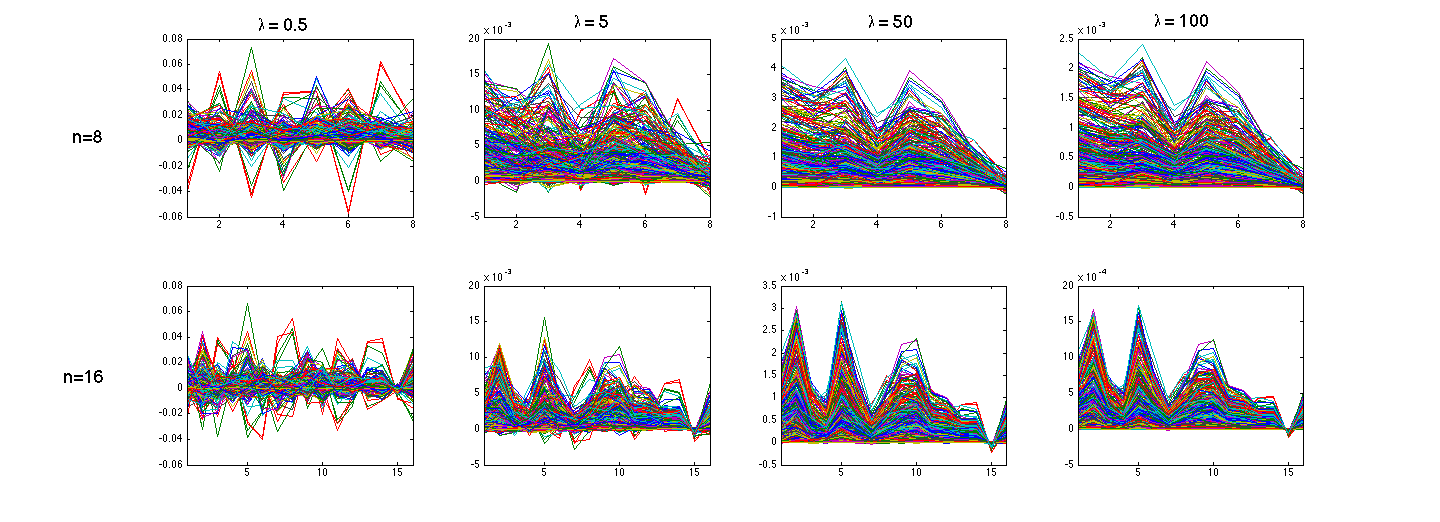}
	%}
	\vspace{-15px}
	\caption{The parameterization $C$ is $D\times n$ matrix. Each plot has $D$ lines, and each line is a plot for values progression of a row in $C$. At large values of $\lambda$ the parameterization is smooth enough to capture large dynamics in the visual unit.%For $n=16$ the parameterization holds more informatoin than for $n=8$
	}
	\label{Fig:comapare-lambda-n}
\end{figure*}
%We have a set of images sequences representing different visual units. 
\subsection{Framework description}
\label{sec:framework}
Manifold Kernel Partial Least Squares (MKPLS) framework is proposed in \cite{Bakry2013}. For convenience, we briefly describe the mathematical model and the framework pipelines here.%in this section.
%

%\subsection{Mathematical model}
%\textbf{Mathematically,}
Let us denote the $k$-th sequence by  $S_k= \{ \vect{x}^k_i \in \mathbb{R}^D, i=1\cdots n_k \}$, 
where the image $\vect{x}^k_i\in \mathbb{R}^D$. Let $y_k$ represents the class labels for the $k$-th sequence. For the particular case of speech recognition and speaker identification, $y_k\in \{c_1, \cdots, c_K\} \times \{p_1, \cdots p_L \}$.  Here $c_i$ is the speech label, and $p_j$ is the performer identity. 
Let $\mathcal{M}_k\subset \mathbb{R}^D$ is a low-dimensional manifold connects the images of sequence $S_k$.  The basic assumption is that all these manifolds ($\mathcal{M}_k \forall k$) are topologically equivalent, however each of them has different geometry in $\mathbb{R}^D$. This assumption is stated clearly in \cite{Bakry2013}. %In other words, all these manifolds are deformed instances of each others. This assumption is fairly met in the domain of activity recognition. For example, periodic locomotive activities intuitively lie on one-dimensional closed manifolds, and hence topologically equivalent.  For instance, sequence of features representing a Viseme, starting from a neutral pose and reaching a peak pose, lies on a one-dimensional manifold (curve) in the feature space. 

%%%%%%%%%%%%%%%%%%%%%%%%%%%%%%%%%%%%%%%%
%We use this framework for evaluating the performance of all manifold kernels that we discuss in this paper. 

 %Fourth, this framework uses kernel based dimensionality reduction technique which gives the ability to investigate different manifold kernels. 
%
%
In principal, \textbf{MKPLS pipeline} has three phases: individual manifold parameterization, latent space embedding and finally inference/classification. 
Consider the manifold $\mathcal{M}_k$ connects the $n_k$ frames of specific Visual Unit (VU).  Assuming, that we have a unified manifold $\mathcal{U}$.
%
%MKPLS represents each visual unit in manifold formate, and 
The result of \textbf{individual manifold parameterization} is a single representation for VU which holds the topological deformation of $\mathcal{M}_k$ with respect to $\mathcal{U}$. Any further processing is done based on these parameterizations. % This step minimizes the effect of the temporal variations between visual units. 

The manifold $\mathcal{M}_k$ is represented by a parameterization $\mathbf{C}_k$ with respect to a set of basis $\{\psi_1,\psi_2,\cdots,\psi_n\}$. These basis are a nonlinear function of points on $\mathcal{U}$. We use Gaussian Radial Basis Function (Gaussian-RBF):
$\psi_{i}(\vect{z})=exp(\sigma\left\|\vect{z}-\vect{w}_i\right\|)$
%\begin{equation}
% \label{eq:gaussian-rbf-psi}
% \psi_{i}(\vect{z})=exp(\sigma\left\|\vect{z}-\vect{w}_i\right\|)
% \end{equation}
where $\vect{w}_i; i =1\cdots n$ are fixed points on $\mathcal{U}$.
%
%[K(\vect{z},\vect{w}_1), \cdots, K(\vect{z},\vect{w}_n)]$ . 
The goal is to find a regression function $\mathcal{\gamma}(t) = \mathbf{C}_k^\top \Psi(t)$ which minimize the objective function  %$\mathcal{\gamma}(t) $ is chosen to minimizes a regularized loss functional in the form 
\begin{equation}
\label{eq:gamma}
    \sum_i^{n^k} \left\|\vect{x}^k_{i} - \gamma^k(\mathbf{z}^k_i)\right\|^2 + \lambda \; \Omega[\gamma^k], 
\end{equation}
where $\left\|\cdot\right\| $ is the Euclidean norm,  $\Psi(t)=[\psi_1(\vect{z}_t),\psi_2(\vect{z}_t),\cdots,\psi_n(\vect{z}_t)]^\top$, $\Omega$ is a regularization function that enforces the smoothness in the learned function, and $\lambda$ is the regularizer. %that balances between fitting the training data and smoothing the learned function. When $\lambda\to 0$, the regression function over-fits the training data. 
Representer theory helps to find closed form for $\mathbf{C}$ as
\begin{equation}
\label{eqn:coeff_matrix}
\mathbf{C}_k^\top  = (\mathbf{A}_k^\top\mathbf{A}_k+\lambda \mathbf{G})^{-1}\mathbf{A}_k^\top\mathbf{X}_k^\top, 
\end{equation}
where $\mathbf{A}_k$ is an $n_k \times n$ matrix with $\mathbf{A}_{(ij)}= exp(\sigma\left\|\vect{z}_i-\vect{w}_j\right\|)$ %, i=1, \cdots,n_k,  j=1 \cdots n$ 
and $\mathbf{G}$ is an $ n \times n$ matrix with $\mathbf{G}_{(ij)}=exp(\sigma\left\|\vect{w}_i-\vect{w}_j\right\|)$%;  i,j=1 \cdots n$
. $\mathbf{X}_k$ is the $n_k \times D$ data matrix for UV$_k$. The details of learning the manifold parameterization can be found in \cite{Bakry2013}. %Solution for $\mathbf{C}$ is guaranteed under certain conditions on the basis functions \cite{Poggio1990:GRBF}. In this paper, we use Gaussian Radial Basis Function (Gaussian-RBF) for the kernel $K(\cdot,\cdot)$.

The choice of $\lambda$ and $n$ is crucial for better performance. Figure~\ref{Fig:comapare-lambda-n} shows the trade-off between value of $\lambda$ and $n$. This choice depends upon the application. In this work, we need to capture the smooth dynamics in the visual units. Therefore, we choose $\lambda=50$. It is clear that $n=16$ expose more variations than with $n=8$. More information can be useful in some cases and can be more confusing in others. Threrefore, in Section~\ref{sec:experments}, we show both configurations.

%%%%%%%%%%%%%%%%%%%%%%%%%%%%%%%%%%%%%%%%%%%%%%%%%%%%%%%
%
%%%%%%%%%%%%%%%%%%%%%%%%%%%%%%%%%%%%%%%%%%%%%%%%%%%%%%%

In \textbf{Latent space embedding}, kernel partial least squares (KPLS) \cite{Rosipal2002:KPLS}  is adopted for embedding the parameterizations $\{\mathbf{C}_k, k=1\cdots N\}$ into a low-dimensional latent space $\mathbb{R}^m$, as $\{\mathbf{t}_k\in \mathbb{R}^m, k=1\cdots N\}$. KPLS is supervised method, so it uses the set of labels $\{y_k, k=1\cdots N\}$ for the embedding. Supervised embedding guarantees to acheive the most concise and informative low-dimensional latent space embedding.
For any manifold $\mathcal{M}_\nu$, represented by its parameterization $\mathbf{C}_\nu$, the corresponding embedded point can be computed by
\begin{equation}
\label{eqn:embedding}
\mathbf{t_\nu = \mathbf{v}_\nu R}.
\end{equation}
Where the projection matrix $\mathbf{R}$ is learned from KPLS algorithm, and $\mathbf{v}_\nu=K(C_\nu,.)\in \mathbb{R}^N$. $K$ measures the similarity between $C_\nu$ and all training manifold parameterizations $\{\mathbf{C}_k, k=1\cdots N\}$.
The choice of kernel $K$ and its computation is discussed in detailed in Section\ref{sec:kernels}.
Because we solve two problems, speech recognition and speaker identification, we learn one $\mathbf{R}$ for each task. We learn $\mathbf{R}_c$ based on speech labels, and we learn $\mathbf{R}_p$ for speaker identification with subject labels. As a result, for each $\mathbf{v}_\nu$, we get two embedding in two latent spaces: $\mathbf{t}_\nu^c=\mathbf{v}_\nu \mathbf{R}_c$ in the speech latent space and $\mathbf{t}_\nu^p=\mathbf{v}_\nu \mathbf{R}_p$ in the speaker latent space.

Finally in \textbf{manifold classification}, given a latent space embedding $\mathbf{t}_\nu$, MKPLS uses several techniques to classify it such as 
Regression for classification (RfC) \cite{Bakry2013}, Support vector machines (SVM) and K-nearest neighbor (KNN). In the latent space of speech, we want to infer the speech label ($c$) while in the speaker space, we need to infer the subject label ($p$). 

%Use regression results of KPLS
%\[
%%L_\nu = t_\nu T^\top\mathbf{L}
%\hat{y}_\nu = \mathbf{t}_\nu \mathbf{T}^\top\mathbf{y}
%\]
%where $t_\nu$ is computed from Eq~\ref{eqn:embedding} (details are in \cite{Bakry2013}). 
%Second choice is Support vector machines (SVM). A one-vs-all SVM classifier is learned using the set of training points $\{(\mathbf{t}_k,y_k)\in \mathbb{R}^m\times \mathbb{R};\ k=1\cdots N\}$.

%
%\section{Single frame classification}
%\section{Toward continuous speech recognition}
%In the section \ref{sec:classify}, we assumed that we have all frames of the new sequence $\mathcal{S}^\nu$ are avaialble. Toward continuous speech recognition, we have to consider the case of classifying the phrase/speaker from solely single or couple of frames. 
%
%For doing that, we need to have generative model. Recipe of generative model is 1) descriptor for single frame 2) some measurement for how sure we are that the image in hand belong to certain sequence 3) classifier for he choosen sequence to know the word and the speaker of this speech video.
%We can do that using sampling method. So we sample from both the 
% % % % % % % % % % % % % % % % % % % % % % % % % % % % % % % % % %

%%%%%%%%%%%%%%%%%%%%%%%%%%%%%%%%%%%%%%%%
\begin{table*} %[!ht]
\caption{Subject Semi-dependent speech recognition on \textbf{OuluVs} database}
%\vspace{2mm}
\setlength\tabcolsep{4pt}{ %\scriptsize
\begin{tabular}{|l|ccccccc|ccccccc|}

\hline
& \multicolumn{7}{|c|}{$n=8$ and $_{1\times 2}\mathbf{LBP}^{u_2}_{1-8\times 8}$}& \multicolumn{7}{|c|}{$n=16$ and $_{1\times 1}$ $\mathbf{LBP}^{u_2}_{1-8\times 8}$}\\
\hline
%$m$& \textbf{SVM} &\textbf{RfC}&\textbf{SVM} &\textbf{RfC}\\
\multicolumn{1}{|r|}{$m=$}&$10$&$30$&$50$&$80$&$100$&$130$&$200$&$10$&$30$&$50$&$80$&$100$&$130$&$200$\\

\hline\hline

Cosine&$62.19$&$78.13$&$81.72$&$81.41$&$81.72$&$82.03$&$81.09$&$58.28$&$77.19$&$79.22$&$79.53$&$79.22$&$79.84$&$79.53$\\

Euclid&$61.25$&$79.06$&$79.38$&$79.53$&$79.84$&$80.16$&$78.91$&$56.72$&$75.16$&$75.00$&$75.63$&$75.94$&$75.31$&$74.69$\\

EditDist&$62.50$&$75.63$&$66.72$&$43.44$&$22.81$&$21.41$&$22.34$&$59.53$&$70.16$&$61.25$&$41.25$&$35.00$&$24.69$&$15.62$\\

Frechet&$29.53$&$27.81$&$25.47$&$17.34$&$15.97$&$14.53$&$11.88$&$ $&$ $&$ $&$ $&$ $&$ $&$ $\\

Grassm&$28.91$&$37.34$&$41.87$&$42.19$&$39.69$&$37.66$&$31.88$&$24.38$&$26.41$&$29.53$&$28.44$&$26.25$&$25.63$&$25.63$\\

GrassmCC&$28.91$&$37.34$&$41.87$&$42.19$&$39.84$&$37.34$&$27.81$&$24.53$&$26.41$&$29.53$&$28.44$&$25.94$&$25.31$&$21.72$\\
GrassmDiff&&&&&&&&$28.13$&$35.00$&$37.81$&$39.17$&$37.29$&$31.67$&$25.63$\\
\hline
\end{tabular}
}
\label{fig:ouluvs-SSD}
\vspace{-10px}
\end{table*}
\section{Manifold-to-manifold Kernels}
\label{sec:kernels}
The parameterization, extracted out of the first phase of MKPLS, holds the dynamics in each video which encodes speech-related information along with speaker-related information. %Figure\ref{eq:gamma} shows that the parameterizations contrast the similarity due to visual features dominates the similarity between speaker than the similarity between 
Because MKPLS uses kernel-based approach for %embedding the parameterizations into the corresponding latent space
dimensionality reduction, the kernel choice is critical for achieving the best performance. In this section, we investigate several types of kernels.

MKPLS claims that, to define manifold-to-manifold kernel, it suffices to define it in the parameterization space,i.e, % described in Section~\ref{sec:framework}, a kernel in the space of manifolds can be defined as a kernel between their parameterizations
%\begin{equation}
%\label{eq:kernel_general}
$ K_{manifold}( \mathcal{M}_i,\mathcal{M}_j) \doteq K(\mathbf{C}_i,\mathbf{C}_j).
$ % \end{equation}
Therefore, we need to define kernels over the space of parameterizations, which consequently,  measure the similarity between manifolds in terms of their geometric deformation from the common manifold representation. 
MKPLS gives us the ability to plugin any valid kernel. In this section, we investigate several choice of kernels: matrix-based kernels, curve-based kernels and subspace-based kernels.
%distance metric such as cosine-distance and Euclidean distance, and subspace-distance kernels such as Grassmannian kernels % and Remainnian kernels,
%and curve-to-curve based kernels such as cosine-similarity kernels~\cite{Bakry2013}, Fr\'{e}chet-distance kernels and Edit-distance kernels.
%%%%%%%%%%%%%%%%%%%%%%%%%%%%%%%%%%%%%%%%
\subsection{Matrix-based kernels}
Since the dimensionality of all parameterizations is unique, we can measure the similarity between them by measuring the similarity between the corresponding column. This is the idea behind the matrix-based kernels. %when comparing the parameterization from the matrix form point of view, we measure  
\subsubsection{Cosine-similarity kernel (Cosine)} 
We can measure the similarity between columns using cosine the angle between them. As a result, the overall similarity between two parameterizations is the sum over all colmn-wise similarities. Therefore, the cosine-manifold kernel can be defined as
\begin{equation}
\label{eq:cos_kernel}
   K_{\cos}(\mathbf{C}_i,\mathbf{C}_j) = \frac{tr(\mathbf{C}_i \mathbf{C}_j^\top)^2} { ||\mathbf{C}_i||_F ||\mathbf{C}_j ||_F},
\end{equation}
where $\left\| \cdot \right\|_F$ is matrix Frobenius norm.
\subsubsection{Euclidean-distance kernel (Eculid)}
In this kernel, we measure the Euclidean distance between the $i$-th column in parameterization $C_1$ ($u_{1i}$) and its corresponding column in parameterization $C_2$ ($u_{2i}$). Hence, the overall matrix kernel $\delta=\sum_{i=1}^{n}||u_i-v_i||_2^2$, and the matrix similarity is
\begin{equation}
\label{eq:kernel_distance}
K(\mathbf{C}_1,\mathbf{C}_2) = \exp(-\omega \delta) 
\end{equation}
where $\omega$ is a normalization factor.
For $K(.,.)$ to be valid kernel, it needs to be symmetric positive definite matrix (SPD). The exponential function takes care of the positive definiteness part. For the symmetry, the used distance measure should be metric, which is satisfied for Euclidean distance case.
%
%This kernels is the exponential of the Euclidean distance 
%%%%%%%%%%%%%%%%%%%%%%%%%%%%%%%%%%%%%%%%
%\subsection{Isotonic Distance Kernels}
\subsection{Curve-based Kernels}
In this category, we consider the columns of the parameterization matrix as points in $\mathbb{R}^D$, and the matrix defines a curve connecting those points. The matching between columns should obey the ordering. % From this point of view, %we need to measure distances between the curves. %Therefore,
%we use a temporally restricted, but more general distance metric for parameterizations. Not like the column-wsie distance, there is not restriction that 
%for any two matrices, each column can be matched with any column in the other matrix, without creating cross matching. Not creating cross matching
This means that if two columns $i,j$ from the first matrix match the columns $u,v$ from other matrix respectively, the  $u \leq v$ iff $i< j$.
%and % Since it does not restrict the comparison to the corresponding columns only but it obey the temporal progression between column. 
%This kind of metrics is the closest to curve-to-curve comparison and time series analysis.
For each of the following distances, the parameterization kernel is computed using Eqn~\ref{eq:kernel_distance}.

\subsubsection{Fr\'{e}chet-distance Kernel (Frechet)}
Fr\'{e}chet distance is a known metric to measure the distance between two curves, that takes into account the location and ordering of the points along the curves. Here, we use discrete Fr\'{e}chet distance, also known as coupling distance, in which we assume that the curves are piece-wise linear. % In this case, we consider each parameterization as piece-wise linear curve in $R^{D}$. 
The basic idea that, each point in one curve is matched with its closest point on the other curve, and the distance $d$ will be the maximum Euclidean distance between each two matched points.
At the end, not all points are matched between the two curves. 

\subsubsection{Edit-distance kernel (EditDist)}
The idea of this metric is similar to minimum edit distance between strings. Two main difference between EditDist and Frechet algorithms: in EditDist the overall distance is the sum of distance between all matches while Frechet takes the maximum of all matches, and EditDist considers unmatched points as being matched with the origin while Frechet ignores the unmatched points. %For each two parameterizations, We use a simple dynamic programming algorithm to measure the minimum costs we need to pay to match the two parameterizations. % In this case, we obey the temporal order of the columns.
Emperically, we found that the best column-wise distance, in both EditDist and Frechet, is the Euclidean distance.
%%%%%%%%%%%%%%%%%%%%%%%%%%%%%%%%%%%%%%%%

\begin{table*} [!htb]
\caption{Speaker Indepenent - speech recogniton Accuracy on \textbf{OuluVs} database}
%\vspace{2mm}
\setlength\tabcolsep{4pt}{ %\scriptsize
\begin{tabular}{|l|ccccccc|ccccccc|}

\hline
& \multicolumn{7}{|c|}{$n=8$ and $_{1\times 2}\mathbf{LBP}^{u_2}_{1-8\times 8}$}& \multicolumn{7}{|c|}{$n=16$ and $_{1\times 1}$ $\mathbf{LBP}^{u_2}_{1-8\times 8}$}\\
\hline
%$m$& \textbf{SVM} &\textbf{RfC}&\textbf{SVM} &\textbf{RfC}\\
\multicolumn{1}{|r|}{$m=$}&$10$&$30$&$50$&$80$&$100$&$130$&$200$&$10$&$30$&$50$&$80$&$100$&$130$&$200$\\
\hline\hline
Cosine&$49.22$&$50.00$&$51.41$&$49.84$&$50.00$&$49.84$&$50.00$&$50$&$51.88$&$49.06$&$49.06$&$50.31$&$50.00$&$50.63$\\
Euclid&$48.59$&$55.16$&$55.78$&$55.00$&$55.47$&$55.47$&$55.00$&$48.44$&$61.25$&$58.44$&$57.50$&$57.81$&$58.75$&$57.19$\\
EditDist&$47.97$&$50.94$&$41.25$&$26.09$&$21.72$&$18.13$&$16.41$&$48.44$&$48.44$&$43.75$&$37.81$&$31.56$&$21.87$&$16.25$\\
Frechet&$11.25$&$13.28$&$12.50$&$12.34$&$13.75$&$13.44$&$13.75$&$ $&$ $&$ $&$ $&$ $&$ $&$ $\\
Grassm&$29.84$&$30.31$&$32.34$&$31.09$&$28.59$&$24.84$&$31.56$&$22.19$&$23.75$&$25.00$&$19.69$&$21.25$&$19.06$&$15.31$\\

GrassmCC&$29.69$&$30.31$&$32.34$&$31.09$&$28.75$&$23.91$&$23.28$&$22.19$&$23.75$&$25.00$&$19.69$&$21.25$&$19.37$&$17.19$\\
\hline
\end{tabular}
}
\label{fig:ouluvs-SI}
\vspace{-10px}
\end{table*}
%%%%%%%%%%%%%%%%%%%%%%%%%%%%%%%%%%%%%%%%
\subsection{Subspace-based Kernel}
Each parameterization $\mathbf{C}_k$ represents $n$-dimensional subspace in $\mathbb{R}^D$. Therefore, we can use subspace-to-space metric to measure the similarity in parameterization space.
This gives the most general comparison between matrices. Because it considers the subspace spanned by the columns of each parameterization without encoding any ordering.
\subsubsection{Grassmannian kernel}
Every coefficient matrix $\mathbf{C}_k$ is $D \times d$. Since $D\gg d$, hence $\mathbf{C}_i$ represent $d$ dimensional subspace in $\mathbb{R}^D$. Therefore, the matrix $\mathbf{C}$ belongs to Grassmannian manifold  $G_{D,d}$. For more details about Grassmannian manifolds, the reader is referred to \cite{Edelman1998}. 

There are several approaches for measuring the similarity on Grassmannian manifold, we use the one defined in \cite{Harandi-cvpr2011}.
%Since the mapping matrices are not orthonormal, we use Gram-Schmidt algorithm [] for orthonormalize the coefficient matrix. By orthonormalize, we mean to get the another matrix which columns are orthonormal and covers the column space of the orgianl matrix. 
\begin{equation}
\label{eq:grassm}
\mathbf{K}_{ij} = a_1\mathbf{K}^{cc}_{ij}+a_2\mathbf{K}^{proj}_{ij}
\end{equation}
Where  $\mathbf{K}^{proj}_{ij}$, $\mathbf{K}^{cc}_{ij}$ are the projection kernel and the canonical correlation kernel respectively, and $a_1,a_2$ are weighting constants.
The projection kernel is defined by
%\begin{equation}
$\mathbf{K}^{proj}_{ij} = \left\|\mathbf{\Pi}_i^\top\mathbf{\Pi}_j\right\|^2_F
$ %\end{equation}
where %$\left\|.\right\|_F$ is the Frobenius norm, and 
$\mathbf{\Pi}_k$ is the orthogonal version of the coefficient matrix $\mathbf{C}_k$, computed by Gram-Schmidt orthogonalization algorithm. 
The canonical correlation kernel is defined by
\begin{equation}
\label{eq:grasCC}
\mathbf{K}^{cc}_{ij}=\max_{a_p\in span\{\mathbf{\Pi}_i\}}\max_{b_q\in span\{\mathbf{\Pi}_j\}}a_p^\top b_q
\end{equation}
Subject to $a_p^\top a_q=b_p^\top b_q=1$ if $p=q$, and $0$ otherwise.
We use two Grassmannian kernels: \textbf{Grassm} defined by Eq~\ref{eq:grassm} and  \textbf{GrassmCC} defined by Eq~\ref{eq:grasCC}.

Since Grassmannian distance does not consider the ordering of the parameterization columns, we can encode some temporal information by using the parameterization of difference of the input features. We denote this experiment by \textbf{GrassmDiff}.% In this case, we consider this difference as a different featis This is one advantages of 
%%%%%%%%% Derivative %%%%%%%%%%%%
%\paragraph{Encoding Derivative}
%
%\[
%\frac{\partial \psi(\theta)}{\partial \theta}   = \frac{\partial \psi(z)}{\partial \theta} \frac{\partial z}{\partial \theta}
%\]
%, where $z = [cos(\theta) \,sin(\theta)]^\top$ and $\psi(z)$ defined (in Eq $5$ in the paper) as
%
%and its Jacobian matrix 
%\[\frac{\partial \psi(z)}{\partial z}  = 2\sigma\left[\frac{(z-w_1)}{\left\|z-w_1\right\|}e^{\sigma\left\|z-w_1\right\|},\frac{(z-w_2)}{\left\|z-w_2\right\|}e^{\sigma\left\|z-w_2\right\|}\cdots \frac{(z-w_n)}{\left\|z-w_n\right\|}e^{\sigma\left\|z-w_n\right\|}\right]\]
%, and
%\[\frac{\partial z}{\partial \theta} = [-sin(\theta) \,cos(\theta)]^T\]

%%%%%%%%%%%%%%%%%%%%%%%%%%%%%%%%%%%%%%%%
%\subsubsection{Remanian Kernel}
% % % % % % % % % % % % % % % % % % % % % % % % % % % % % % % % % %

\begin{table*} [!tbh]
\caption{Speaker Identification Accuracy on \textbf{OuluVs} database}
%\vspace{2mm}
\setlength\tabcolsep{4pt}{ %\scriptsize
\begin{tabular}{|l|ccccccc|ccccccc|}

\hline
& \multicolumn{7}{|c|}{$n=8$ and $_{1\times 2}\mathbf{LBP}^{u_2}_{1-8\times 8}$}& \multicolumn{7}{|c|}{$n=16$ and $_{1\times 1}$ $\mathbf{LBP}^{u_2}_{1-8\times 8}$}\\
\hline
%$m$& \textbf{SVM} &\textbf{RfC}&\textbf{SVM} &\textbf{RfC}\\
\multicolumn{1}{|r|}{$m=$}&$10$&$30$&$50$&$80$&$100$&$130$&$200$&$10$&$30$&$50$&$80$&$100$&$130$&$200$\\
\hline\hline
Cosine&$93.91$&$99.69$&$99.69$&$99.69$&$99.69$&$99.69$&$99.69$&$92.66$&$99.69$&$99.69$&$99.69$&$99.69$&$99.69$&$99.69$\\

Euclid&$93.75$&$99.53$&$99.53$&$99.53$&$99.53$&$99.53$&$99.53$&$93.91$&$99.53$&$99.53$&$99.53$&$99.69$&$99.53$&$99.53$\\

EditDist&$94.22$&$99.53$&$99.53$&$99.37$&$99.06$&$71.88$&$69.69$&$92.81$&$99.69$&$99.53$&$99.53$&$99.53$&$99.69$&$99.53$\\

Frechet&$83.91$&$95.16$&$89.06$&$75.94$&$64.84$&$50.63$&$17.97$&$88.13$&$96.09$&$87.81$&$62.81$&$27.66$&$27.66$&$27.66$\\

Grassm&$84.69$&$99.38$&$99.37$&$99.53$&$99.06$&$98.75$&$97.81$&$92.19$&$99.22$&$99.06$&$98.91$&$98.44$&$98.33$&$97.56$\\

GrassmCC&$84.69$&$99.38$&$99.37$&$99.53$&$99.06$&$98.75$&$97.81$&$92.19$&$99.22$&$99.06$&$98.91$&$98.44$&$98.33$&$95.94$\\
\hline
\end{tabular}
}
\label{fig:ouluvs-speaker}
\vspace{-10px}
\end{table*}
\section{Experiments}
\label{sec:experments}
\subsection{Databases}
\label{sec:databases}

We apply MKPLS for OuluVs database \cite{Zhao2009}. OuluVs has ten different everyday phrases. Each phrase is uttered by $20$ subjects up to five times. We use the same test protocoal used in \cite{Bakry2013}. The frame rate was set to $25$ fps. The dataset  contains sequence of images for mouth area with average resolution of $120 \times 60$ pixels. %This database is less constrained than AVLetters, so that limited rotation and shift was allowed in the recording time. %Figure~\ref{fig:ouluvs_examples}(\textbf{a}). %slightly shiftedIn this area, you can find some mouthes slightly rotated and shifted. 
%Not all sequences are perfectly segmented, so that, some sequences have few frames with partial-mouth (Figure~\ref{fig:ouluvs_examples}(\textbf{b})) or non-mouth frames (Figure~\ref{fig:ouluvs_examples}(\textbf{c})). 
%Some of the outlier sequences (that contain very few mouth/partial-mouth frames) are excluded from the experiment. Consequently, we exclude four speakers with very few sequences remaining (\textit{P004}, \textit{P005}, \textit{P010} and \textit{P016}).
%
%\begin{figure}
%\begin{tabular}{c c c}
%\includegraphics[width=0.4\columnwidth]{figs/ouluvs_complete}&
%\includegraphics[width=0.2\columnwidth]{figs/ouluvs_partial}&
%\includegraphics[width=0.2\columnwidth]{figs/ouluvs_nonmouth}\\
%(\textbf{a}) & (\textbf{b})& (\textbf{c})\\
%\end{tabular}
%\caption{OuluVs: (\textbf{a}) Regular frames, (\textbf{b}) Partial mouth area frames, (\textbf{b}) Non-mouth area frames.}
%\label{fig:ouluvs_examples}
%\vspace{-10px}
%\end{figure}
$\mathbf{LBP}$ \cite{Ojala2002} visual features is extacted from images. Two feature configurations have been used: the first configuration is $\mathbf{LBP}_{1:8\times 8}$ with $n=16$ ($16$ basis for Gaussian-RBF) and the second one is
$_{1\times 2}\mathbf{LBP}_{1:8\times 8}^{u_2}$ with $n=8$.
%
% The results is reported in terms of two of them: single cell eight-resolutions ($\mathbf{LBP}_{1:8\times 8}$) and $3\times 4$ cell-grid  with four-resolutions ($_{3\times 4}\mathbf{LBP}_{1:4\times 8}^{u_2}$). For more details about \textbf{LBP}, reader is referred to .
%
%The feature configurations used on this database is  ($\mathbf{LBP}_{1:8\times 8}$) and 
%($_{1\times 2}\mathbf{LBP}_{1:8\times 8}^{u_2}$).
%
We also apply MKPLS for AVLetters database~\cite{Matthews2002-avletters} which has ten subjects. Each speaker repeats every English letter ($A\cdots Z$) exactly three times, with a total of $780$ video sequences. The speaker was requested to start and end utterance of every letter in a neutral state (mouth closed). We apply single feature configuration: $3\times 4$ cell-grid  with four-resolutions LBP features ($_{3\times 4}\mathbf{LBP}_{1:4\times 8}^{u_2}$) with $n=8$.

In all experiments, the recognition rate is measured as the ratio between the correctly recognized clips and the total number of clips.
%\vspace{-12px}

\subsection{Experimental Results}
In this section, we present the empirical results of MKPLS when plugined with each one of the kernels described in Section~\ref{sec:kernels}. To give real comparison between the proposed kernels, we need to explore different parameters of MKPLS pipeline:
The number ($n$) of Gaussain-RBF basis $\psi$ that we use to learn the individual manifold parameterization, we show results for $n={8,16}$.
%Parameter affects the generation of the manifold parameterization which is the number of temporal centers $n$. 
The dimensionlaity of the manifold latent space, we use $m = {10,30,50,80,100,130,200}$, which cover wide range of the possible values.

Especially for Grassmann-based kernels~\ref{sec:kernels}, we show the affect of using the parameterization of the LBP of the images itself vs LBP of the images concatenated with parameterization of the discrete difference between those images.
%This section is organized by experiment protocoals and within each experiment we compare using different kernels.

\subsubsection{Visual speech recognition}
Two test protocols has been adopted for visual speech recognition: Speaker Independent (SI) and Speaker Semi-Dependent (SSD) as defined in ~\cite{Bakry2013}.
%, we had one more protocol "speaker dependent", but we skipped it here, because we need to measure the ability of the framework to balance between the speaker and speech variations. For each configuration, we are going to show the result of the set of kernels discribed in Section~\ref{sec:kernels}.

\textit{Speaker Semi-Dependent VSR (\textbf{SSD}):} Here we test on one repeat of the available videos and train based on the remaining repeats for the same subjects. In this configuration all subjects and phrases are presented in the training set. %The challenge here is to classify the phrase correctly regardless the user identity.
Table~\ref{fig:ouluvs-SSD} show the SSD speech recognition accuracy for OuluVs database with the two feature configurations. Table~\ref{fig:avletters-SSD} shows the result of matrix-based kernels and curve-based kernels applied avletters.
\begin{table} [ht]
\caption{SSD speech recognition on \textbf{AvLetters}}
%\vspace{2mm}
\setlength\tabcolsep{2pt}{ \small
\begin{tabular}{|l|ccccccc|}
\hline
& \multicolumn{7}{|c|}{$n=8$ and $_{3\times 4}\mathbf{LBP}_{1:4\times 8}^{u_2}$}\\
\hline
%$m$& \textbf{SVM} &\textbf{RfC}&\textbf{SVM} &\textbf{RfC}\\
\multicolumn{1}{|r|}{$m=$}&$10$&$30$&$50$&$80$&$100$&$130$&$200$\\
\hline\hline
Cosine&$50.77$&$56.67$&$60.77$&$62.82$&$63.85$&$64.49$&$63.85$\\
Euclid&$51.41$&$56.41$&$60.38$&$64.49$&$65.13$&$64.52$&$64.74$\\
EditDist&$51.41$&$56.54$&$60.51$&$64.36$&$65.13$&$65.00$&$64.74$\\
Frechet&$23.59$&$34.62$&$36.28$&$34.49$&$35.64$&$34.49$&$29.23$\\
\hline
\end{tabular}
}
\label{fig:avletters-SSD}
\vspace{-10px}
\end{table}

\textit{Speaker Independent VSR (\textbf{SI}):} the challenge here is %to recognize the uttered phrase, independent of the speaker. By this configuration. 
to show the scalability of the model, mean how far the model can recognize the spoken phrase based on the dynamics even if the speaker is not seen before in the training set. In this experiment, we use one-speaker-out. Table~\ref{fig:ouluvs-SI} show the SI speech recognition accuracy for OuluVs for the two configurations.
\subsubsection{Speaker Identification (\textbf{SpId}):}
The goal in this experiment is to find the speaker within the register set of users. The challenge is to find the speaker from the limited available information in the mouth area. 
%Moreover, we want to prove that although the manifold parameterization encodes mainly the geometric deformation from the unified manifold to the original data manifold, parameterization also hold speaker-related information. 
We take one repetition out for testing, and train over all other repetitions.
Table~\ref{fig:ouluvs-speaker} shows the speaker identification accuracy when applied to OuluVs for the two test configurations.

\subsection{Discussion}
From the numbers, we can clearly see the superiority of both techniques of Matrix-based kernels (Cosine and Euclid) in all test and features configurations. For $m=10$, EditDist gives the best results for SSD-speech recognitoin and Speaker Identification. %EditDist is a direct generalization to matrix-based EditDist gives good results but still less thanCosine and Euclid.
GrassDiff gives slightly better resutls than Grassm abd GrassCC, since it encodes temporal information. Frechet kernel proves failure in this application. For Avletters, EditDist and Euclid give the best recognition rate.

\section{Conclusion}
\label{sec:conclusion}
%We proposed a framework that utilized the homeomorphic manifold analysis and KPLS for manifold classification. We tackled two related classification problems speaker identification and speech recognition. 
%We use supervised latent low-dimensional space embedding for solving the simultaneous multi-factor classification problem.
%We presented three different configurations of lipreading speaker independent, speaker semi-dependent and speaker dependent. The results show that our approach outperform in the semi-dependent setting which we consider the most realistic configuration and perform well in the other two settings.
We investigated the kernel choice for the middle phase of MKPLS framework. We explored three kind of manifold kernels: matrix-based kernel, curve-based kernel and subspace-based kernel. We compare the kernels based on different parameter configuration for MKPLS. The experiments shows that using parameterization-to-parameterization kernel can delegate manifold-to-manifold kernel. The results shows the superiority of the matrix-based kernel for visual speech recogniton. For speaker identification tasks, all kernels gives perfect results.
\vspace{6pt}
%%%%%%%%%%%%%%% Bibs

%\newpage
%

{ %\small
\bibliographystyle{IEEEtran}
\bibliography{library}

% Generated by IEEEtran.bst, version: 1.14 (2015/08/26)
\begin{thebibliography}{10}
\providecommand{\url}[1]{#1}
\csname url@samestyle\endcsname
\providecommand{\newblock}{\relax}
\providecommand{\bibinfo}[2]{#2}
\providecommand{\BIBentrySTDinterwordspacing}{\spaceskip=0pt\relax}
\providecommand{\BIBentryALTinterwordstretchfactor}{4}
\providecommand{\BIBentryALTinterwordspacing}{\spaceskip=\fontdimen2\font plus
\BIBentryALTinterwordstretchfactor\fontdimen3\font minus
  \fontdimen4\font\relax}
\providecommand{\BIBforeignlanguage}[2]{{%
\expandafter\ifx\csname l@#1\endcsname\relax
\typeout{** WARNING: IEEEtran.bst: No hyphenation pattern has been}%
\typeout{** loaded for the language `#1'. Using the pattern for}%
\typeout{** the default language instead.}%
\else
\language=\csname l@#1\endcsname
\fi
#2}}
\providecommand{\BIBdecl}{\relax}
\BIBdecl

\bibitem{Potamianos2004}
G.~Potamianos and C.~Neti, ``{Audio-visual automatic speech recognition: An
  overview},'' \emph{Issues in Visual and Audio-Visual Speech Processing}.

\bibitem{McGurk1976}
H.~McGurk and J.~MacDonald, ``{Hearing lips and seeing voices},''
  \emph{Nature}, vol. 264, no. 23 December, pp. 746--748, 1976.

\bibitem{Shiell2008}
D.~Shiell and L.~Terry, ``{Audio-Visual and Visual-Only Speech and Speaker
  Recognition: Issues about Theory, System Design, and Implementation},''
  \emph{Visual speech recognition: lip segmentation and mapping}.

\bibitem{Luettin1996}
J.~Luettin, N.~Thacker, and S.~Beet, ``{Speaker identification by
  lipreading},'' \emph{International Conference on Spoken Language Processing},
  pp. 1--4.

\bibitem{Sanderson2004}
C.~Sanderson and K.~Paliwal, ``{Identity verification using speech and face
  information},'' \emph{Digital Signal Processing}, no.~5, pp. 449--480, Sep.

\bibitem{Rabiner1989}
L.~Rabiner, ``{A tutorial on hidden Markov models and selected applications in
  speech recognition},'' \emph{Proceedings of the IEEE}, no.~2, pp. 257--286.

\bibitem{Matthews2002-avletters}
I.~Matthews and T.~Cootes, ``{Extraction of visual features for lipreading},''
  \emph{PAMI}, vol.~24, no.~2, pp. 198--213, 2002.

\bibitem{Saenko2005}
K.~Saenko and K.~Livescu, ``{Visual speech recognition with loosely
  synchronized feature streams},'' \emph{IEEE International Conference on
  Computer Vision}.

\bibitem{Zhou2010}
Z.~Zhou and G.~Zhao, ``{Lipreading: A Graph Embedding Approach},'' \emph{ICPR},
  pp. 523--526, 2010.

\bibitem{Zhou2011}
Z.~Zhou, G.~Zhao, and M.~Pietikainen, ``{Towards a practical lipreading
  system},'' \emph{Computer Vision and Pattern Recognition}, 2011.

\bibitem{Zhao2009}
G.~Zhao, ``{Lipreading with local spatiotemporal descriptors},'' \emph{IEEE
  Transactions on Multimedia}, pp. 1--11, 2009.

\bibitem{Shaikh2010}
A.~Shaikh, D.~Kumar, and W.~Yau, ``{Lip Reading using Optical Flow and Support
  Vector Machines},'' \emph{IEEE International Congress on Image and Signal
  Processing}, vol.~1, pp. 327--330, Oct. 2010.

\bibitem{Bakry2013}
A.~Bakry and A.~Elgammal, ``{MKPLS: Manifold Kernel Partial Least Squares for
  Lipreading and Speaker Identification},'' \emph{Computer Vision and Pattern
  Recognition}, 2013.

\bibitem{Rosipal2002:KPLS}
R.~Rosipal and L.~Trejo, ``{Kernel partial least squares regression in
  reproducing kernel hilbert space},'' \emph{The Journal of Machine Learning
  Research}, pp. 97--123.

\bibitem{Edelman1998}
A.~Edelman, T.~a. Arias, and S.~T. Smith, ``{The Geometry of Algorithms with
  Orthogonality Constraints},'' \emph{SIAM Journal on Matrix Analysis and
  Applications}, no.~2, pp. 303--353, Jan.

\bibitem{Harandi-cvpr2011}
M.~Harandi and C.~Sanderson, ``{Graph embedding discriminant analysis on
  Grassmannian manifolds for improved image set matching},'' \emph{IEEE
  Conference on Computer Vision and Pattern Recognition}, pp. 2705--2712, Jun.

\bibitem{Ojala2002}
T.~Ojala, ``{Multiresolution gray-scale and rotation invariant texture
  classification with local binary patterns},'' \emph{PAMI}, no.~7, pp.
  971--987.

\end{thebibliography}
}

\end{document}